\documentclass[10.5pt,compsoc]{BD}
\usepackage{graphicx}
\usepackage{footmisc}
\usepackage{subfigure}
\usepackage{url}
\usepackage{multirow}
\usepackage[noadjust]{cite}
\usepackage{amsmath,amsthm}
\usepackage{amssymb,amsfonts}
\usepackage{booktabs}
\usepackage{color}
\usepackage{ccaption}
\usepackage{booktabs}
\usepackage{float}
\usepackage{fancyhdr}
\usepackage{caption}
\usepackage{xcolor,stfloats}
\usepackage{comment}
\graphicspath{{figures/}}
\usepackage{cuted}  
\usepackage{captionhack}
\usepackage{epstopdf}

\usepackage{svg}
\usepackage{lipsum}
\usepackage{booktabs}
\usepackage{amsmath} 
\usepackage{tcolorbox}
\usepackage{multirow}
\usepackage{url} 

\usepackage{algorithm}
\usepackage{algpseudocode}
\usepackage{array}
\usepackage{textcomp}
\usepackage{stfloats}
\usepackage{url}
\usepackage{verbatim}
\usepackage{cite}

\headevenname{\normalsize{\textbf{\emph{xxxxxxxxxxxxxxx,} xxxxxxx}} 20xx, x(x): xxx-xxx}%
\headoddname{{\sf Pengfei Liu et al.:}\quad {\textbf{\emph{Enhanced Drug-drug Interaction Prediction Using Adaptive Knowledge Integration}}}}%

\newtheoremstyle{mystyle}{0pt}{0pt}{\normalfont}{1em}{\bf}{}{1em}{}
\theoremstyle{mystyle}
\renewcommand\figurename{Fig.~}

\newcommand{\nop}[1]{}

\makeatletter
\renewcommand{\@biblabel}[1]{[#1]\hfill}
\makeatother

\begin{document}

\thispagestyle{empty}


\clearpage

\hyphenpenalty=50000

\makeatletter
\newcommand\mysmall{\@setfontsize\mysmall{7}{9.5}}

\newenvironment{tablehere}
  {\def\@captype{table}}
  {}
\newenvironment{figurehere}
  {\def\@captype{figure}}
  {}

\thispagestyle{plain}%
\thispagestyle{empty}%

\let\temp\footnote
{}

\vskip .2mm\noindent
{\normalsize\textbf{\scalebox{1}[1.0]{\makebox[5.6cm][s]{%
{V\hfill o\hfill l\hfill u\hfill m\hfill%
e\hspace{0.356em}x,\hspace{0.356em}N\hfill u\hfill%
m\hfill b\hfill e\hfill r\hspace{0.356em}x,\hspace{0.356em}%
x\hfill x\hfill x\hfill%
x\hfill x\hfill x\hfill x \hspace{0.356em}2\hfill0\hfill x\hfill x}}}}}\\

\begin{strip}
{\center
{\LARGE\textbf{
Enhanced Drug-drug Interaction Prediction Using Adaptive Knowledge Integration}}
\vskip 9mm}

{\center {\sf \large
Pengfei Liu, Jun Tao, Zhixiang Ren$^*$
}
\vskip 5mm}

\centering{
\begin{tabular}{p{160mm}}

{\normalsize
\linespread{1.6667} %
\noindent
\bf{Abstract:} {\sf
Drug-drug interaction event (DDIE) prediction is crucial for preventing adverse reactions and ensuring optimal therapeutic outcomes.
However, existing methods often face challenges with imbalanced datasets, complex interaction mechanisms, and poor generalization to unknown drug combinations.  
To address these challenges, we propose a knowledge augmentation framework that adaptively infuses prior drug knowledge into a large language model (LLM).
This framework utilizes reinforcement learning techniques to facilitate adaptive knowledge extraction and synthesis, thereby efficiently optimizing the strategy space to enhance the accuracy of LLMs for DDIE predictions.
As a result of few-shot learning, we achieved a notable improvement compared to the baseline.
This approach establishes an effective framework for scientific knowledge learning for DDIE predictions.
}
\vskip 4mm
\noindent
{\bf Key words:} {\sf Drug-drug interaction event; Large language model; Knowledge prior; Reinforcement learning
}}

\end{tabular}
}
\vskip 6mm

\vskip -3mm
\small\end{strip}

\thispagestyle{plain}%
\thispagestyle{empty}%
\makeatother
\pagestyle{tstheadings}

\begin{figure}[b]
\vskip -6mm
\begin{tabular}{p{44mm}}
\toprule\\
\end{tabular}
\vskip -4.5mm
\noindent
\setlength{\tabcolsep}{1pt}
\begin{tabular}{p{1.5mm}p{79.5mm}}

$\bullet$& Pengfei Liu is with Pengcheng Laboratory, Shenzhen 518000, and also with School of Computer Science and Engineering, Sun Yat-sen University, Guangzhou 510006, China.
E-mail: liupf7@mail2.sysu.edu.cn.\\
$\bullet$& Jun Tao is with the School of Computer Science and Engineering, Sun Yat-sen University, Guangzhou 510006, China.
E-mail: taoj23@mail.sysu.edu.cn.\\
$\bullet$& Zhixiang Ren is with Pengcheng Laboratory, Shenzhen 518000, China.
E-mail: jason.zhixiang.ren@outlook.com.\\
$\sf{*}$& To whom correspondence should be addressed. \\
\end{tabular}
\end{figure}\large

\vspace{3.5mm}
\section{Introduction}
Drug-drug interactions (DDIs) occur when the effects of one drug are altered by the presence of another, potentially leading to enhanced or diminished efficacy or even adverse reactions.
The DDIs can produce different responses in pharmacy, pharmacokinetics, and pharmacodynamics, making them complex to study and often only verifiable through experimental methods.
Drug-drug interaction events (DDIEs) refer to these specific interaction patterns where the interaction between drugs affects their performance and safety.
Beyond merely identifying whether an interaction exists, precise recognition of DDIEs can help healthcare professionals avoid potential adverse effects and tailor safer and more effective medication regimens for individual patients.
However, traditional experimental methods \cite{safdari2016computerized} for studying DDIs are costly and sometimes unfeasible, limiting their practical application in identifying these interactions.

The integration of artificial intelligence (AI) into DDI prediction \cite{zhang2023application}\cite{vilar2012drug}leverages the expanding volume of pharmaceutical data to enhance both accuracy and efficiency.
The INDI \cite{gottlieb2012indi} model utilizes logistic regression classifiers to predict DDIs.
The DeepDDI \cite{ryu2018deep} model, along with its dataset, has significantly advanced the field by enabling predictions of DDIE through the deep neural network (DNN).
Despite these advancements, most existing AI methods still face considerable challenges.
The scarcity and imbalance of high-quality DDI data limit the training and effectiveness of models.
The complexity and diversity of drug interaction mechanisms complicate accurate predictions.
Additionally, current models often lack the capability to generalize effectively to unknown drug combinations, restricting their practical application.

The GPT-4 \cite{achiam2023gpt} has fueled the popularity of large language models (LLMs) and spurred their application across various industries, particularly in AI for science research.
For scientific language modeling \cite{liu2024scientific}, specialized models like Galactica \cite{taylor2022galactica} focus on scientific domains.
In biochemistry, models such as MolT5 \cite{edwards2022translation}, BioT5+ \cite{pei2024biot5+}, and ChemLLM \cite{zhang2024chemllm} can be fine-tuned on DDI datasets to enhance their cross-domain capabilities.
These models use training data such as drug molecules' Simplified Molecular Input Line Entry System (SMILES) strings \cite{weininger1988smiles}, molecular fingerprints, and drug descriptions to improve both generalizability and accuracy in DDI predictions.

\begin{figure*}[t!]
\centering
\includegraphics[width=\textwidth]{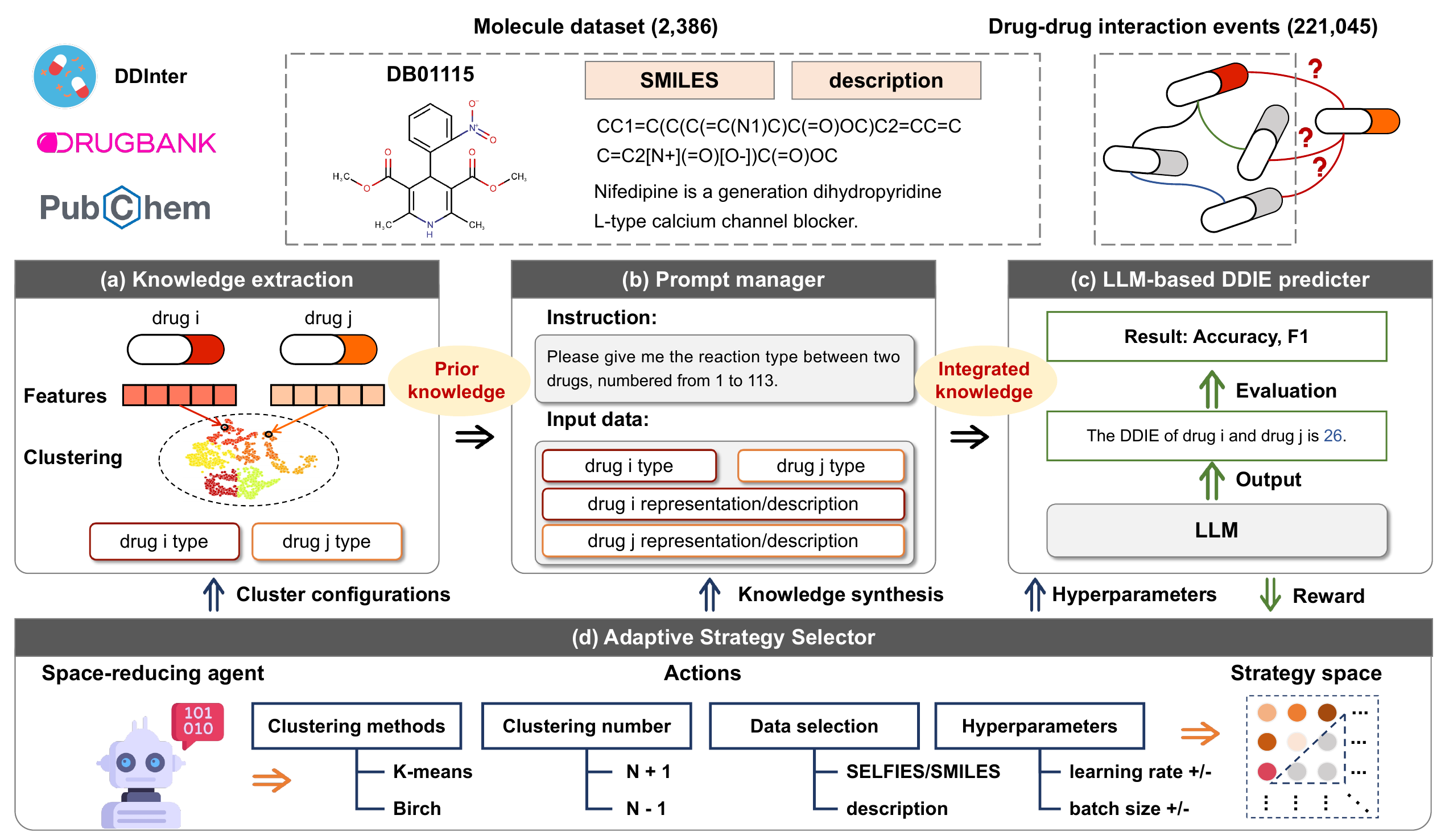}
\caption{\textbf{Overview of the framework}. 
Leveraging the information of drug molecules to enhance the accuracy of DDIE prediction in few-shot scenarios.
(a) Knowledge Extraction classifies drug types from features using specific clustering methods and counts.
(b) Prompt Manager uses drug types as prior knowledge and combines them with molecular representations or descriptions to form prompts.
(c) LLM-based DDIE Predictor predicts DDIEs, with the performance feedback serving as a reward to the Adaptive Strategy Selector.
(d) Adaptive Strategy Selector develops strategies based on clustering methods, knowledge synthesis strategies, and LLM hyperparameters, exploring the strategy space to iteratively select optimal states.
}
\label{fig: overview}
\end{figure*}

Despite these advances, the limited availability of public drug information and known DDIEs challenges LLMs in predicting interactions with new drugs.
Scientific prior knowledge can enhance model performance with limited data, typically through data knowledge priors and model structure priors.
However, these approaches have not yet been implemented in the DDI domain.
Therefore, we can focus on two key issues: 
\textbf{(1) Can prior knowledge improve the performance of LLMs for DDI prediction under conditions of limited data?
(2) If so, given the vast knowledge in the pharmaceutical field, which strategy of knowledge augmentation would yield the best result?}

To tackle these challenges, we propose a knowledge augmentation framework that integrates reinforcement learning (RL) to refine the knowledge within constrained datasets, thereby improving the performance of DDIE predictions.
Specifically, our approach employs clustering methods to categorize drugs, which are merged with different drug representations such as SELFIES \cite{krenn2020self} or textual descriptions as input data.
Moreover, our framework uses RL to adaptively determine the most suitable clustering method, the number of clusters, the drug modality, and the hyperparameters of LLM.
This process assesses strategies to optimize the knowledge integration tailored to the dynamics of drug interactions within the 864-size strategy space.
In summary, our main contributions are the following:
\begin{itemize}
    \item \textbf{Knowledge-prior DDIE prediction:} To our knowledge, this is the first method to improve DDIE predictions in LLMs by incorporating drug knowledge priors.
    \item \textbf{Adaptive knowledge integration:} We develop an end-to-end model using RL that adaptively optimizes the integration patterns of molecular knowledge, effectively narrowing the exploration space.
    \item \textbf{Few-shot performance enhancement:}
    By integrating prior knowledge, the few-shot learning capabilities of LLMs achieve an improvement in the F1-score over the baseline, depending on the data split.
\end{itemize}

The organization of this paper is as follows:
Section~\ref{sec:rel} provides a comprehensive review of related studies, setting the foundation for our proposed approach.
Section~\ref{sec:met} details our methodology, including the data sources, models, and strategies employed.
Section~\ref{sec:eva} focuses on the evaluation of our methods, featuring an ablation study to underscore the significant findings.
Section~\ref{sec:dis} examines the strengths and limitations of our approach and proposes directions for future research.
Finally, Section~\ref{sec:con} concludes the paper by summarizing our primary contributions.
The data and code can be accessed at \url{https://github.com/AI-HPC-Research-Team/Drug_drug_interaction_with_LLM}.

\section{Related work}
\label{sec:rel}
In this section, we explore various models for DDI predictions and delve into the specialized application of LLMs, particularly in the molecular domain.
Additionally, we discuss the innovative use of knowledge extraction and priors, and the integration of RL for data filtering and integration.
These discussions lay the foundation for our proposed approach.
\subsection{Drug-drug interaction}

The datasets such as DrugBank \cite{wishart2018drugbank} and TWOSIDES \cite{tatonetti2012data} provide extensive information on DDI triples and drug side effects.
Additionally, researchers have developed databases tailored specifically for drug-drug interactions, such as DDInter \cite{xiong2022ddinter}, a comprehensive and open-access database dedicated to DDIs.
For a more granular understanding of interaction events, the DeepDDI dataset features 86 distinct DDIE labels.
DeepDDI 2 \cite{kim2023enhanced} expands the classification to include a comprehensive 113 types of drug interactions.
DeepDDI employs drug features and a DNN for DDIE prediction.
Similarly, graph models integrate drug features processed by DNNs with drug topology using graph neural networks (GNNs) to generate representative embeddings for DDI prediction, such as DM-DDI \cite{kang2022multitype}, SSI-DDI \cite{nyamabo2021ssi} and DSN-DDI \cite{li2023dsn}.
For textual drug representations like SMILES, TBPM-DDIE \cite{shao2022tbpm} extracts latent vectors with a Transformer \cite{vaswani2017attention}, incorporating both structural and semantic drug information, which are then utilized for DDIE classification.
These methodologies focus on drug representations and training classifiers for multi-class tasks, which typically suffer from \textbf{poor generalizability}.

\subsection{Large language models}
Transformer-based models such as BERT \cite{devlin2018bert}, GPT \cite{radford2019language}, and T5 \cite{raffel2020exploring} have demonstrated exceptional capabilities in comprehending and generating human language.
In molecular sciences, there are generative LLMs such as MolT5 \cite{edwards2022translation}, ChemT5 \cite{christofidellis2023unifying}, BioT5 \cite{pei2023biot5}, and MolXPT \cite{liu2023molxpt}, as well as encoding models like the single-modal Chem-BERTa \cite{chithrananda2020chemberta} and multi-modal models such as MolFM \cite{luo2023molfm} and GIT-Mol \cite{liu2024git}, designed to handle complex chemical information.
As for link prediction tasks, such as drug-target interaction (DTI) and DDI,
the method \cite{simon2024prediction} that utilizes BERT for analyzing protein sequences and drug compounds has proven effective in facilitating robust DTI predictions.
BioT5+ \cite{pei2024biot5+}, a SELFIES-based pre-trained model, incorporates the IUPAC \cite{long1983limit} names of molecules during its pre-training phase, enabling it to execute molecular and protein-related tasks.
The utilization of LLMs for DDI and DTI predictions often involves the challenge of designing effective prompts for drug pairs.
For example, TextDDI \cite{zhu2023learning} leverages RoBERTa-Base \cite{liu2019roberta} to process drug descriptions, thereby enhancing the accuracy of DDI predictions.
Nevertheless, the broad and diverse knowledge base concerning drug molecules, including various drug categories and their different representations, presents significant challenges in \textbf{preparing and optimizing prompts for LLMs}.

\subsection{Knowledge priors}
Knowledge priors are established, domain-specific insights that can be effectively integrated into AI models to considerably enhance their analytical and predictive capabilities.
This integration may be implemented within the model's architecture, training strategies, and dataset, enriching the model’s depth of understanding and enhancing its predictive accuracy.
In molecular science, the incorporation of knowledge priors \cite{kuang2024impact}, including molecular structures, chemical properties, and reaction mechanisms, plays a crucial role in augmenting the predictive precision of AI models.
The KPGT \cite{li2023knowledge} framework exemplifies this approach by employing a graph transformer for molecular graphs, coupled with a knowledge-guided pretraining strategy designed to comprehend both the structural and semantic aspects of molecules.
Similarly, 
The KANO \cite{fang2023knowledge} model utilizes an element-oriented knowledge graph as a prior, learning fundamental knowledge of functional groups to achieve enriched molecular representations.
In LLMs, the approach \cite{liu2024self} extracts knowledge patterns from chemical reactions and injects these into the LLM, thereby enhancing reaction predictions.
DDIPrompt \cite{wang2024ddiprompt} exemplifies the application of knowledge priors during its pretraining phase by training a GNN model using the molecular structure of drugs and the similarity of interactions between drugs.
While prior research has validated the benefits of incorporating prior knowledge to boost model performance, challenges persist in effectively \textbf{extracting and integrating the knowledge}.

\subsection{Reinforcement learning for data optimization}
RL entails agents learning to make optimal decisions in a dynamic environment by performing actions that maximize cumulative rewards.
This method capitalizes on feedback from the agent's actions, enabling a progressive refinement of its decision-making process to achieve superior outcomes over time.
Tasks related to the organization of existing data can be framed as feature selection or combinatorial optimization challenges.
The method \cite{lee2017automatic} employs RL to optimize sentence selection for text summarization tasks, using the ROUGE-2 score as a reward metric to enhance the quality of summaries.
SemiACO \cite{karimi2023semiaco} defines the feature selection space as a Markov Decision Process, employing a semi-supervised method based on ant colony optimization to enhance feature selection efficiently.
Additionally, TextDDI \cite{zhu2023learning} applies RL to iteratively refine and select optimal drug descriptions for accurate DDI predictions, improving prediction accuracy through continual feedback and adjustments.
Furthermore, these RL-based approaches hold significant potential for \textbf{optimizing the combination of knowledge} within prompts for LLMs.

\section{Methodology}
\label{sec:met}

In this section, we present an overview of our framework, which is built upon existing pre-trained LLMs without modifying their internal architecture.
Our innovation lies in the adaptive integration of prior drug knowledge into the LLM to enhance DDIE prediction under limited data conditions.
Specifically, we utilize clustering methods to extract drug-type knowledge, which is then combined with molecular representations or textual descriptions through a Prompt Manager (PM).
These prompts are then passed into the LLM, offering a more flexible and natural mechanism for knowledge integration compared to graph-based models.
Finally, we apply RL to iteratively optimize the knowledge extraction strategy, drug modality, and hyperparameters, effectively improving the model’s performance in few-shot scenarios.

\subsection{Problem formulation}

The problem of predicting DDIEs inherently involves a \textbf{multiclass classification framework}.
The data includes a set of drugs \( \mathcal{D} \), where each drug \( d \in \mathcal{D} \) is associated with information \( x_d \). The interactions between drugs are represented as pairs \( (d_i, d_j) \) with associated interaction types from a set \( \mathcal{E} \).
The model \( f \) maps pairs of drug features to interaction types, where \( e_{ij} \) represents the type of interaction between drugs \( d_i \) and \( d_j \).
\begin{equation}
f(x_{d_i}, x_{d_j}) \rightarrow e_{ij} \quad \text{for} \quad e_{ij} \in \mathcal{E}.
\end{equation}

Our framework can \textbf{enrich the data with domain-specific knowledge} and integrate it into the LLM to improve DDIE prediction.
We extract drug types \( t_{d} \) as categorical knowledge from feature vectors \( v_{d} \) in drug data \( x_d \):
\begin{equation}
t_{d} = \text{Cluster}(v_{d}),
\end{equation}
where the function Cluster denotes the clustering process that categorizes drugs into types.
This drug type \( t_{d} \) along with its information \( x_d \) are combined and used as input to the LLM:
\begin{equation}
\text{LLM}(t_{d_i}, x_{d_i}, t_{d_j}, x_{d_j}) \rightarrow e_{ij}.
\end{equation}
This approach leverages prior knowledge and drug representations to enhance the prediction capability of the LLM for various DDIEs.

We define a strategy exploration space \(S\) that includes clustering methods \( \mathcal{C}_M \), cluster number \( \mathcal{N} \), data modalities \( \mathcal{M} \), and model hyperparameters \( \mathcal{H} \).
We employ RL to iteratively optimize this exploration space. The space is defined as:
\begin{equation}
S = \{\mathcal{C}_M, \mathcal{N}, \mathcal{M}, \mathcal{H}\},
\end{equation}
\begin{equation}
\{\mathcal{C}_M, \mathcal{N}, \mathcal{M}\}
\rightarrow t_{d}, x_{d}.
\end{equation}

The RL framework dynamically adjusts these parameters based on the performance feedback from the LLM, thus enabling continuous improvement:
\begin{equation}
\text{RL}(S) \rightarrow \text{Optimized}(S).
\end{equation}
We aim to maximize the performance of the LLM, which is evaluated by metrics such as accuracy or F1-score on DDIE prediction.
The framework can adaptively find the optimal strategy that yields better model performance.

\subsection{Data}
\begin{figure*}[t!]
\centering
\includegraphics[width=0.95\textwidth]{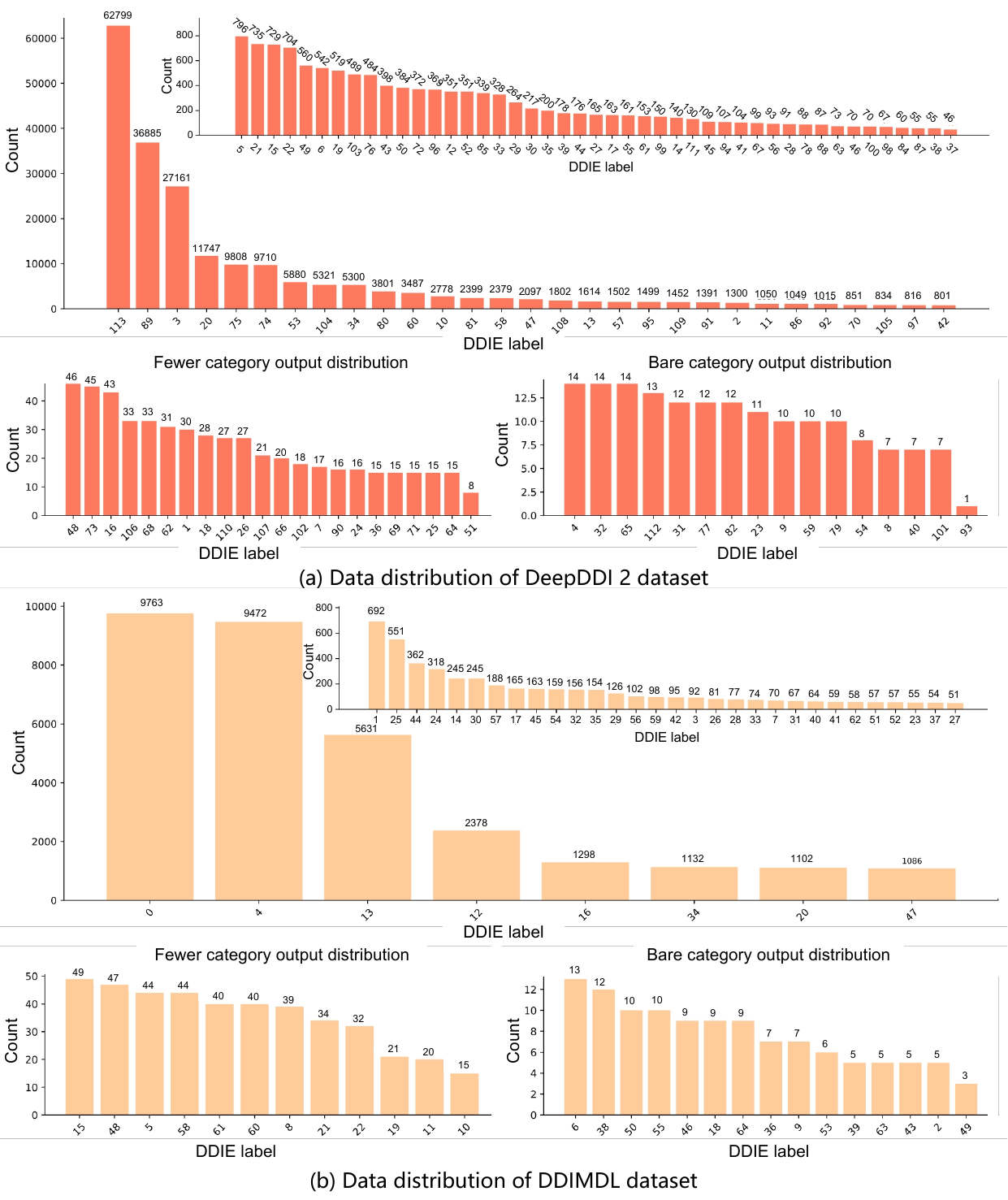}
\caption{\textbf{Dataset analysis.}
(a) and (b) show the frequency distribution of DDIE categories in the DeepDDI 2 and DDIMDL datasets.
Categories are grouped into Common, Few, and Rare based on sample frequency, with the y-axis showing the number of samples and the x-axis showing the category indices.
Note that although some numerical labels may overlap across datasets, the underlying categories differ.}
\label{fig:figure2_clustering}
\end{figure*}

We obtained the drug and DDIE dataset from the DeepDDI 2 \cite{kim2023enhanced}.
This drug dataset includes information on 2,386 drugs, each represented by a 50-dimensional Principal Components Analysis feature vector \( v_d \) and the drug's SMILES strings.
Additionally, we obtained descriptions from DDInter, PubChem \cite{kim2023pubchem}, and DrugBank, which were added to the drug dataset \( \mathcal{D} \).
The DDIE dataset, with a total of 222,127 drug pairs, provides a robust foundation for predicting 113 types of DDIEs.
Similarly, we process the DDIMDL dataset in the same manner, with a total of 567 drugs and 37,137 drug pairs.
It offers a robust basis for predicting 65 types of DDIEs.
The distribution of interaction type categories is shown in Figure \ref{fig:figure2_clustering}.
The interaction types are drawn from a set \( \mathcal{E} \), where the DDIE type \( e_{ij} \) between drug \( d_i \) and drug \( d_j \) belongs to \( \mathcal{E} \).
The importance of addressing few-shot scenarios in our study stems from the frequent occurrence of rare and poorly documented drug interactions in clinical settings, which can pose significant challenges in real-world applications.
Few-shot learning is crucial for developing models that can accurately predict outcomes based on limited examples, thereby enhancing the model's practical utility and reliability when encountering less common drug interactions.

\begin{table}[t!]
\centering
\caption{Distribution of data samples across different interaction frequency categories in training, validation, and test sets.}
\label{tab:frequency_distribution}
\vspace{0.3cm}
\begin{tabular}{>{\centering\arraybackslash}p{0.2\linewidth} >{\centering\arraybackslash}p{0.22\linewidth} >{\centering\arraybackslash}p{0.1\linewidth} >{\centering\arraybackslash}p{0.1\linewidth} >{\centering\arraybackslash}p{0.1\linewidth}}
\toprule
\textbf{Dataset} & \textbf{Frequency} & \textbf{Train} & \textbf{Valid} & \textbf{Test} \\ \midrule
\multirow{3}{*}{DeepDDI 2 \cite{kim2023enhanced}} 
    & Common (50$<$) & 44,126 & 44,113 & 132,110 \\
    & Few (15-50)   & 108    & 128    & 298     \\
    & Rare ($<$15)  & 43     & 34     & 85      \\ \midrule
\multirow{3}{*}{DDIMDL \cite{deng2020multimodal}}
    & Common (50$<$) & 7,331  & 7,308  & 21,958  \\
    & Few (15-50)   & 76     & 86     & 263     \\
    & Rare ($<$15)  & 20     & 33     & 62      \\ \bottomrule
\end{tabular}
\end{table}

For categorization for few-shot learning, the dataset were classified into three hierarchical levels based on the frequency of DDIE: `rare' for occurrences less than 15, `few' for occurrences between 15 and 50, and `common' for more than 50 occurrences, inspired by the approach used in DDIPrompt \cite{wang2024ddiprompt}.
It is crucial for addressing the challenges posed by different frequency categories in real-world scenarios.
Moreover, we removed categories with fewer than two samples and then distributed the remaining samples into training, validation, and test sets at ratios of 2:2:6 to enhance the quality and reliability of the dataset.
This step ensures that our dataset contains those categories that have sufficient data to support meaningful analysis and reliable prediction outcomes.
The training data provides sufficient coverage for all categories, including the `rare' and `few' categories.
The validation set serves to prevent overfitting by providing a mechanism to fine-tune the model parameters before final evaluation.
The test set is reserved for testing to rigorously assess the model’s performance across varied conditions, ensuring robustness and reliability, as detailed in Table \ref{tab:frequency_distribution}.

The data split strategy supports model training and evaluation by ensuring a balance between learning from extensive training data and testing under practical conditions.
Then, we converted all drug SMILES to the SELFIES format to standardize molecular representation.
Finally, the dataset was streamlined to consist of 221,045 samples, prepared for detailed analysis and model training.
In this study, we focused on extracting and processing drug molecular features to ensure that our subsequent analysis, including clustering and visualization, was both robust and interpretable.

In our study, we used 50-dimensional molecular features of drugs from the DeepDDI 2 dataset \cite{kim2023enhanced}.
These features represent a comprehensive set of attributes designed to capture the nuanced interactions and characteristics of drug molecules. 
The 50-dimensional molecular features for each drug were derived using a methodical process to ensure the features were both informative and manageable for predictive modeling.
The chemical fingerprints of each drug were first calculated using extended-connectivity fingerprints of diameter 4 (ECFP4), implemented with the Python package RDKit.
ECFP4 is a widely used technique that encodes the structural information of molecules, capturing the presence of specific substructures within the drug compounds.

\subsection{Knowledge extraction}

The extraction of drug attribute knowledge typically involves dimensionality reduction techniques applied to drug fingerprints. 
In our study, we extract 50-dimensional drug feature representations from the DeepDDI 2 dataset \cite{kim2023enhanced}.
These features, derived from ECFP4 using the RDKit package, encode molecular chemical structures and capture the presence of specific substructures.
Given the high dimensionality of the data, we explored several dimensionality reduction methods, including t-SNE \cite{van2008visualizing}, PCA \cite{mackiewicz1993principal}, and UMAP \cite{mcinnes2018umap} (Supplementary Information II).
Our DDIE prediction task, which involves text generation using an LLM, requires that clustering labels be converted into one-dimensional categorical representations for integration into LLM prompts.
High-dimensional (50D) embeddings cannot be directly utilized as input, necessitating dimensionality reduction to 2D or 1D to facilitate both visualization and the generation of stable, interpretable clustering labels.
This 2D reduction is a core design requirement of our framework.

As detailed in the Supplementary Information IV, t-SNE outperformed PCA and UMAP in generating well-defined clusters, achieving a Silhouette Score of up to 0.497 and a Trimmed Purity of 0.947 when aligned with ATC Level-1 categories.
These results confirm t-SNE’s suitability for capturing pharmacologically relevant structures, supporting its use in our framework.
The resulting 2D features were clustered to generate drug category labels, which we define as prior knowledge in our framework.
These labels, derived from ECFP4 molecular fingerprints, encapsulate the chemical structure characteristics of drugs, providing a foundation for integrating domain-specific knowledge into the LLM for DDIE prediction.
Once the features are transformed into a 2-dimensional space, we perform clustering to uncover the underlying patterns in the drug feature data.
The 50D clustering yielded poor quality (e.g., Silhouette Score -0.046 for K-means, Davies-Bouldin Score 2.54--3.25), confirming that 2D t-SNE clustering produces more stable and meaningful results.
Finally, we chose t-SNE for its superior ability to preserve local data structure, effectively reducing the 50-dimensional feature space to a 2-dimensional representation that facilitates subsequent analysis.

To address the stability of high-dimensional clustering, we also evaluated clustering performance on the original 50D features using K-means \cite{ahmed2020k}, Birch \cite{zhang1996birch}, and Agglomerative \cite{murtagh1983survey} algorithms, as reported in the Supplementary Information III and IV.
The number of clusters is varied between 5 and 20 to explore different granularity levels of drug categorization \cite{xiong2022ddinter}.
This range is selected to reflect the realistic diversity and complexity of drug types encountered in practical scenarios, balancing oversimplification and excessive fragmentation.
The choice of 5 to 20 clusters is deemed appropriate, as it provides sufficient detail without diluting meaningful insights, thereby enhancing the interpretability and applicability of the resulting drug categories.
We define the clustering process as the function \( \text{Cluster} \), which is explicitly defined by the clustering method and the number of clusters. 
This is represented in the following equation:
\begin{equation}
\text{Cluster}(v_d, \mathcal{C}_M, \mathcal{N}) = \{T_1, T_2, \ldots, T_{\mathcal{N}}\},
\end{equation}
where \( v_d \) represents the set of data points to be clustered, \( \mathcal{C}_M \) specifies the clustering algorithm used (e.g., K-means, Birch, Agglomerative), and \( \mathcal{N} \) denotes the number of clusters into which the data is to be divided.
Each \( T_i \) represents a cluster formed from the drug feature data, which is considered as drug types \( t_{d} \) and prior knowledge for the LLM.

Compared to traditional grid search, which exhaustively evaluates all possible parameter combinations, Q-learning offers a more efficient, adaptive search process.
It prioritizes exploring promising regions of the strategy space based on reward feedback, thereby significantly reducing computational cost in a large combinatorial space (864 strategies in our study).

\subsection{Knowledge injection}

We have annotated the drug categories in the drug dataset through knowledge extraction, providing prior knowledge for data preparation.
The drug type is denoted as \( t_{d} \), the context description is denoted as \( s \), the representation is denoted as \( r \) (SMILES or SELFIES representation), and the data modalities \( \mathcal{M} \).
The integrated prompt \( p_{ij} \) can be formulated as:
\begin{equation}
x_{d} = [v_{d}; r_{d}; s_{d}], \mathcal{M} = \{r, s\},
\end{equation}
\begin{equation}
p_{ij} = \text{PM}(\tau,t_{d_i}, t_{d_j}, r_{d_i}, r_{d_j}, s_{d_i}, s_{d_j}),
\end{equation}
where \( \tau \in \mathcal{T} \) denotes a selected prompt template from a predefined set \( \mathcal{T} \) of prompt variations (e.g., direct imperative, question-based, or role-playing styles).

Our prompt engineering strategy aims to optimize LLM performance in DDIE prediction by employing a fixed template with key parameters (e.g., drug category labels, SELFIES, or textual descriptions) replaced to integrate prior knowledge adaptively.
The PM is central to our model's ability to leverage prior knowledge effectively.
It is responsible for structuring the inputs related to drug types, contexts, and representations, ensuring that the model can make informed predictions.
Specifically, the drug type is denoted as \( t_d \), the context description as \( s_d \), and the molecular representation as \( r_d \).
The PM processes these elements by generating prompts that guide the model's understanding and prediction tasks.

Examples of these prompt templates, including variations in tone and guidance style (e.g., direct imperative, question-based, and role-playing), are provided in Supplementary Information V for detailed reference.
These templates enable flexible integration of $t_d$, $s_d$, and $r_d$, allowing the model to adapt to structural (e.g., SELFIES) or contextual knowledge while maintaining simplicity and effectiveness in few-shot scenarios.

The prepared prompts are fed to the LLM, and performance is evaluated using accuracy and F1-score as metrics for the multi-class classification task.
The classification task can be formulated as:
\begin{equation}
\hat{e}_{ij} = \text{LLM}(p_{ij}),
\end{equation}
where \( \hat{e}_{ij} \) is the predicted DDIE type. The results demonstrate the efficacy of our approach for leveraging prior knowledge and advanced language models to predict DDIE.

The hyperparameters of an LLM significantly impact its performance.
Typical hyperparameters include batch size, learning rate, number of layers, hidden units, and dropout rate.
Batch size controls the number of samples processed before updating the model, with larger sizes offering stable gradients and smaller sizes helping to escape local minima.
Learning rate determines the step size during training, with higher rates speeding up convergence but risking overshooting, while lower rates provide precise convergence at the cost of longer training times.
For LLMs, changing the number of layers and hidden units can significantly affect pre-trained model performance, unlike MLPs.
We choose batch size, dropout rate, and learning rate as optimization hyperparameters \( \mathcal{H} \) because they directly influence the balance between training efficiency and model accuracy.
\begin{equation}
\mathcal{H} = \{\text{batch size}, \text{learning rate}\ , \text{dropout rate}\}.
\end{equation}

\subsection{Knowledge combination optimization}

The knowledge combination optimization is a form of input data optimization.
Its purpose is to select, combine, and derive a rich information dataset from existing data, making it easier for the model to understand and learn.
Equation (4) defines the strategy space.
We defined three clustering methods \( \mathcal{C}_M \), 16 clustering numbers \( \mathcal{N} \), two molecular modalities \( \mathcal{M} \), and hyperparameters \( \mathcal{H} \) that contain three options each for batch size and learning rate.
This results in a strategy space comprising 864 strategies, from which we need to find the optimal solution that balances efficiency and performance.

Clustering methods and the number of clusters directly influence prior knowledge.
Different molecular modalities represent the intrinsic properties of molecules, while hyperparameters such as batch size and learning rate affect the model's performance.
Especially in few-shot learning scenarios, these variations can lead to significant differences in model performance.

\begin{algorithm}
\caption{Q-learning for knowledge combination}
\begin{algorithmic}[1]
\State Initialize Q-table with zeros
\State Define the strategy space \( S = \{\mathcal{C}_M, \mathcal{C}_N, \mathcal{M}, \mathcal{H}\} \)
\State Set initial\_state
\State Set best accuracy \(\mathcal{A}_{best}\), best F1-score \(\mathcal{F}_{best}\) to zero
\State Set threshold \(\tau\) maximum iterations without improvement
\State Initialize agent with Q-table and other parameters
\For{each episode}
    \State Set done to False
    \State Set \( c \) to zero
    \While{not done}
        \State \( s \leftarrow \) agent.select\_strategy()
        \State Apply \( s \) to initial\_state, train and evaluate LLM
        \State Obtain accuracy \(\mathcal{A}\) and F1-score \(\mathcal{F}\)
        \State Reward \(\mathcal{R} = (\mathcal{A} - \mathcal{A}_{best}) + (\mathcal{F} - \mathcal{F}_{best})\)
        \State Observe new state \( s' \)
        \State agent.update\_Q(s, \(\mathcal{R}\), s')
        \State \(\mathcal{A}_{best}, \mathcal{F}_{best}, c \leftarrow \begin{cases}
            (\mathcal{A}, \mathcal{F}_{best}, 0) & \text{if } \mathcal{A} > \mathcal{A}_{best} \\
            (\mathcal{A}_{best}, \mathcal{F}, 0) & \text{if } \mathcal{F} > \mathcal{F}_{best} \\
        \end{cases}\)
        \If{\(\mathcal{A} \leq \mathcal{A}_{best} \) and \(\mathcal{F} \leq \mathcal{F}_{best} \)}
            \State \( c = c + 1 \)
        \EndIf    
        \If{ \( c \geq \tau \) }
            \State done = True
        \EndIf
    \EndWhile
\EndFor
\State \Return optimal strategy from Q-table
\end{algorithmic}
\label{alg: q-learning}
\end{algorithm}

To find the optimal strategy within the large strategy space, we implement knowledge combination optimization using RL, with Q-learning being a suitable method.
Q-learning is effective for this task because it enables the exploration and exploitation of various strategies to maximize metrics such as accuracy and F1-score.
While our approach leverages standard Q-learning techniques, it serves as a practical tool to enable adaptive selection of optimal strategies, significantly enhancing efficiency over exhaustive grid search.
As detailed in Section IV, Q-learning reduced the number of evaluations compared to grid search, allowing effective navigation of the 864-strategy space to optimize knowledge integration for DDIE prediction.
Q-learning refines the strategy to optimize the combination of knowledge.

As described in Algorithm \ref{alg: q-learning}, Q-learning for Knowledge Combination Optimization involves initializing a Q-table, defining the strategy space, and setting performance baselines.
Through multiple episodes, it uses an epsilon-greedy policy to select and apply strategies, trains and evaluates the LLM, updates Q-values based on rewards, and terminates if no performance improvement is observed beyond a threshold \(\tau\), ultimately returning the optimal strategy.
For more detailed algorithmic descriptions (including the RL state/action space design, reward formulation, clustering integration, and prompt template evolution), please refer to Supplementary Information VI.

\section{Evaluation and results}
\label{sec:eva}
\begin{figure*}[t!]
\centering
\includegraphics[width=\textwidth]{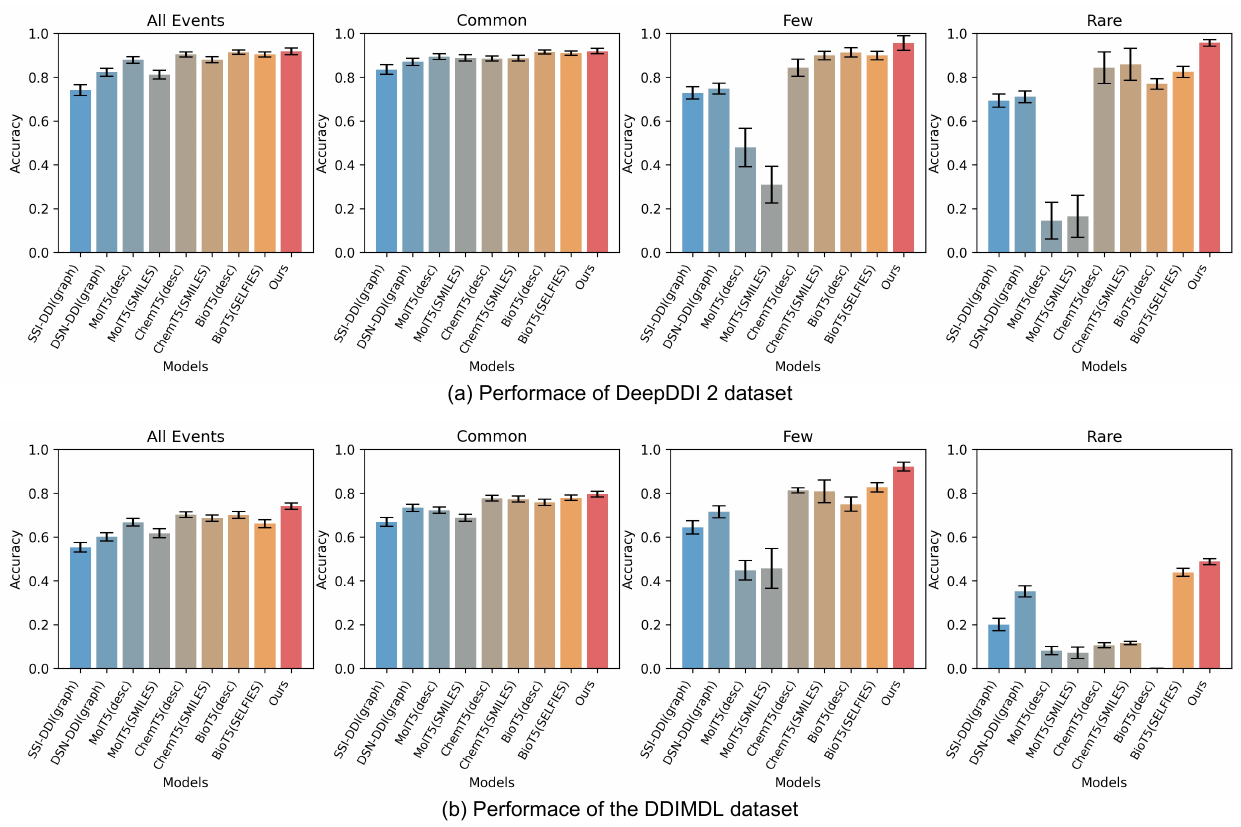}
\caption{\textbf{Accuracy performance of baseline models and ours on DeepDDI 2 (a) and DDIMDL (b) datasets across data splits (All Events, Common, Few, Rare).}}
\label{fig: result}
\end{figure*}

In this chapter, we subject our proposed methodologies to a rigorous evaluation aimed at addressing two fundamental key issues that guide our investigation.
For the first key issue, we implemented drug knowledge extraction through clustering and integrated it with drug molecule representations to enhance the LLM in few-shot learning scenarios, ultimately improving the accuracy of DDIE prediction.
For the second key issue, we developed an adaptive strategy selector based on RL to explore the optimal strategy within a large strategy space, which includes methods of knowledge extraction, knowledge combination modes, and LLM hyperparameters.
The accuracy performance of the models across different data splits is illustrated in Figure \ref{fig: result}, while detailed classification metrics including Accuracy, Precision, F1-score, and Recall are provided in Supplementary Information Table III.

\begin{figure*}[t!]
\centering
\includegraphics[width=0.85\textwidth]{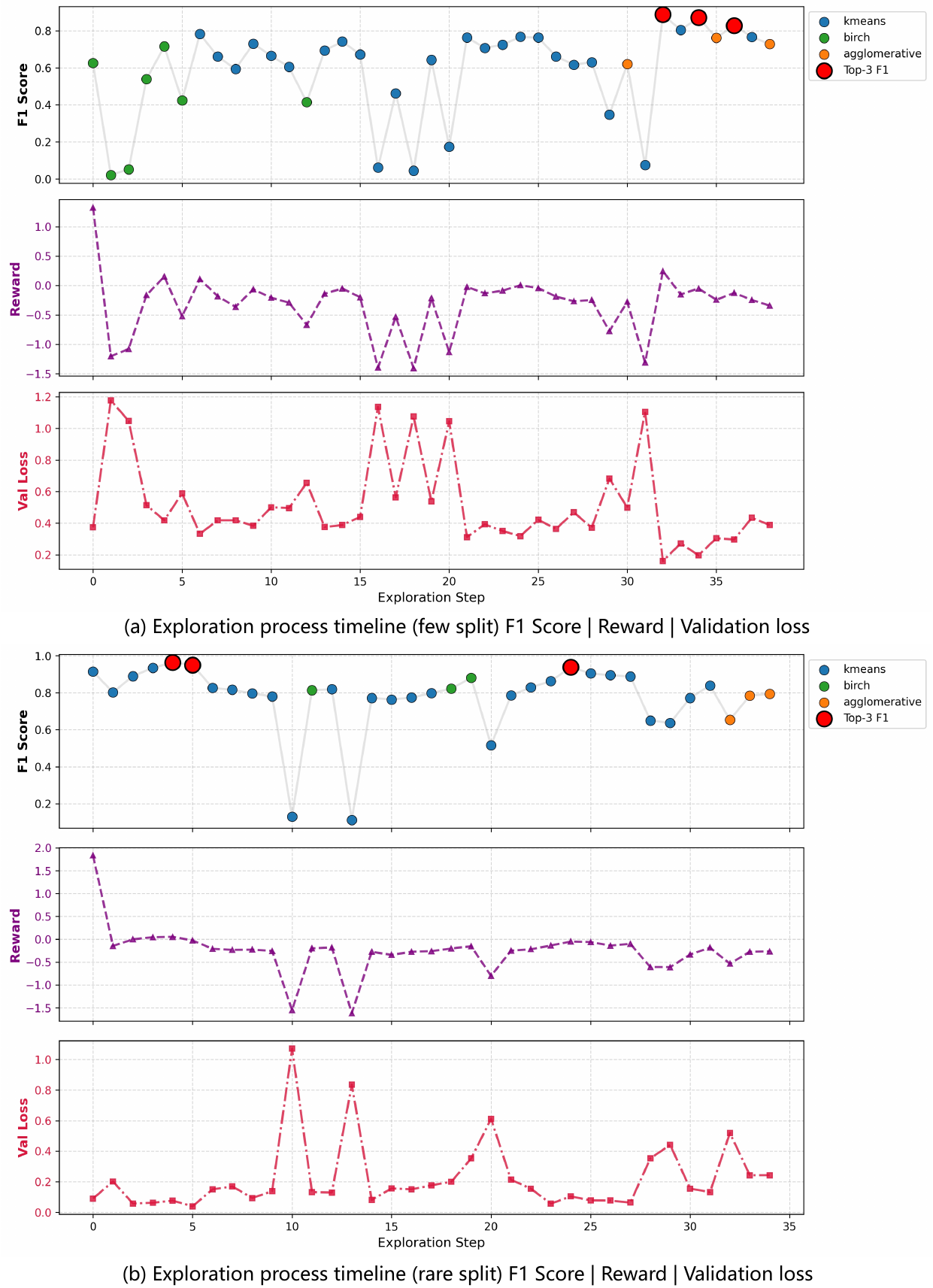}
\caption{\textbf{The exploration process of strategy combinations for the few and rare data splits in DeepDDI 2 dataset}. 
The horizontal axis represents the exploration steps, while the vertical axis shows the F1-score evaluation metric. 
Different colors indicate the corresponding clustering methods.
The top-3 F1 scores in each split are highlighted with red circles. The right axes show the corresponding Reward (purple dashed line) and Validation Loss (crimson dash-dot line).
}
\label{fig: appendix_figure2}
\end{figure*}

\subsection{Experimental setup}

The primary aim of our framework is to enrich the knowledge representation of data, thereby enhancing the LLM's comprehension abilities under limited data conditions.
The dataset is divided by frequency into `rare', `few', and `common' categories, with an additional all-inclusive category, and experiments are conducted across these four divisions.
We select two representative graph-based models (SSI-DDI \cite{nyamabo2021ssi}, DSN-DDI \cite{li2023dsn}) and three LLMs (MolT5 \cite{edwards2022translation}, ChemT5 \cite{christofidellis2023unifying
}, BioT5 \cite{pei2023biot5}) as baseline candidates.
We evaluate all methods using standard classification metrics, including Accuracy, Precision, Recall, and F1-score (Supplementary Information Section VIII).
To ensure a fair comparison and directly address the concern on reproducibility, all baseline LLMs are re-trained from scratch under the same training schedule, identical prompts template, and the same three random seeds (42, 0, 1) as our method.
Baselines use only raw molecular representations (SMILES, SELFIES, or descriptions) with random prompt templates, without any clustering or RL-guided strategies.

Based on the preliminary experimental results, we select ChemT5 as the core model for our framework.
The training process utilizes Tesla V100-SXM2-32GB GPUs with CUDA 11.7 and PyTorch 2.0.0. We employ the AdamW optimizer to enhance training efficiency and performance.
Batch sizes vary from 12 to 24 to suit different dataset divisions, and early stopping is implemented with patience of 2 epochs to prevent overfitting.
Learning rates are set between 5e-4 and 1e-3 to balance fine-tuning speed and model stability.

\subsection{Main results}
To address the two key issues, we can explore the optimal knowledge combination strategy to obtain relatively optimal results.
For the different data splits, particularly the few and rare categories, we configured the number of episodes to be 10 and set the threshold \(\tau\) to 10.
Additionally, we randomly initialized the initial state in Algorithm \ref{alg: q-learning}.
After the training script execution started, we continuously explored the results and eventually obtained over 100 sets of results, accounting for 1/7 of the parameter space.

In our study, we employed a RL approach to efficiently explore the parameter space.
The objective was to identify parameter configurations that yield optimal model performance without exhaustively searching the entire parameter space.
This approach allows the model to learn and adaptively focus on regions of the parameter space that are more likely to improve performance.
As illustrated in Figure \ref{fig: appendix_figure2}, the model's performance evolves progressively as the exploration steps advance.
Notably, there are specific points in the exploration where the performance remains consistently high, indicating the discovery of promising parameter configurations.
In addition, the reward signal and validation loss are tracked at each step.
Although the search process sometimes exhibits large fluctuations in both reward and validation loss—reflecting the exploration of less effective configurations—these instabilities are rapidly corrected in subsequent steps.
The trajectories ultimately converge and stabilise once high-performing regions of the search space are identified, with further exploration terminated by the predefined patience mechanism.

In Figure \ref{fig: result}, we compare the few-shot performance of several fine-tuned LLMs across different data splits.
ChemT5 consistently outperforms BioT5 and MolT5, likely because the latter models tend to generate outputs unrelated to DDIE categories in few-shot settings.
It is especially noticeable in our open-ended generation task, where models are not strictly constrained and must learn to produce the correct class index based on the prompt.
Figure \ref{fig: appendix_figure2} presents a detailed comparison of clustering strategies, input modalities, and hyperparameter settings.
The results show that there is no single best combination across all conditions.
The effectiveness of the strategy is influenced by data characteristics such as interaction frequency and molecular representation.
Throughout the exploration, we also observed that some parameter configurations yielded performance metrics lower than the baseline.
It highlights the importance of systematically exploring the parameter space to avoid suboptimal settings and ensure the model is tuned to achieve its optimal performance.
Therefore, this exploration process is crucial for determining optimal parameters, as it prevents the adoption of configurations that could negatively affect the model's effectiveness.

\begin{figure*}[t!]
\centering
\includegraphics[width=\linewidth]{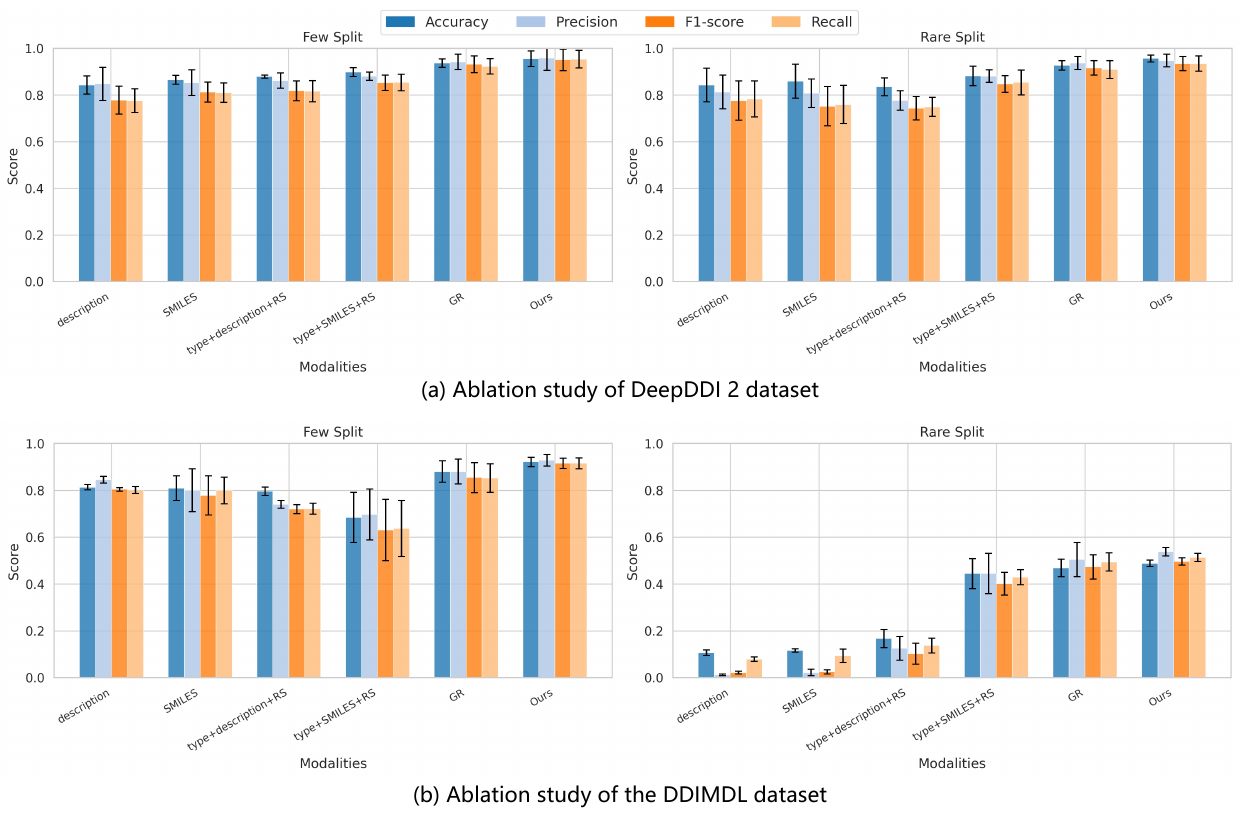}
\caption{Ablation study on DeepDDI 2 (a) and DDIMDL (b) datasets for Few and Rare splits. 
``GR'' denotes coarse grid search, ``RS'' denotes random search, and ``Ours'' is the proposed RL-based searcher.
Error bars indicate standard deviation over three runs.}
\label{fig:ablation}
\end{figure*}

While we did not explore the entire parameter space, our approach successfully identified parameter sets that achieved high model performance.
This method, with its proven success in narrowing down the search to configurations that significantly enhance the model's predictive accuracy, instills confidence in the utility of RL in optimizing model training.

\subsection{Ablation study}
To ensure a fair comparison of different search paradigms under the same computational budget, we design four progressively sophisticated ablation settings:

\begin{itemize}
  \item \textbf{Baseline}: ChemT5 fine-tuned on single-modality input (SMILES or molecular description) without clustering or any search strategy.
  \item \textbf{Clustering + Random Search (RS)}: random sampling of a configuration from the full discrete strategy space of 864 candidates (comprising modality selection, cluster numbers, learning rate, batch size, and prompt–knowledge fusion patterns).
  \item \textbf{Clustering + Grid Search (GR)}: a coarse grid covering approximately 1/4 of the full space.
  \item \textbf{Clustering + RL Search (Ours)}: our reinforcement-learning-based searcher
\end{itemize}
Note that prompt templates are not part of the search space; multiple manually designed prompts are used only during final evaluation to demonstrate generalization.

As shown in Figure~\ref{fig:ablation}, despite our RL-based method exploring only 1/4 to 1/2 of the search space examined by grid search, it consistently discovers superior configurations.
Specifically, it outperforms the best grid-searched strategy by 1.5--4.0\% on DeepDDI 2 and achieves particularly large gains on the more challenging DDIMDL Rare split.
This demonstrates that RL can navigate the high-dimensional discrete space more efficiently and escape poor local regions that uniform or grid-based sampling frequently falls into.
More detailed ablation results are shown in the Supplementary Information (Section VIII, Table IV).

\begin{figure*}[t!]
\centering
\includegraphics[width=\linewidth]{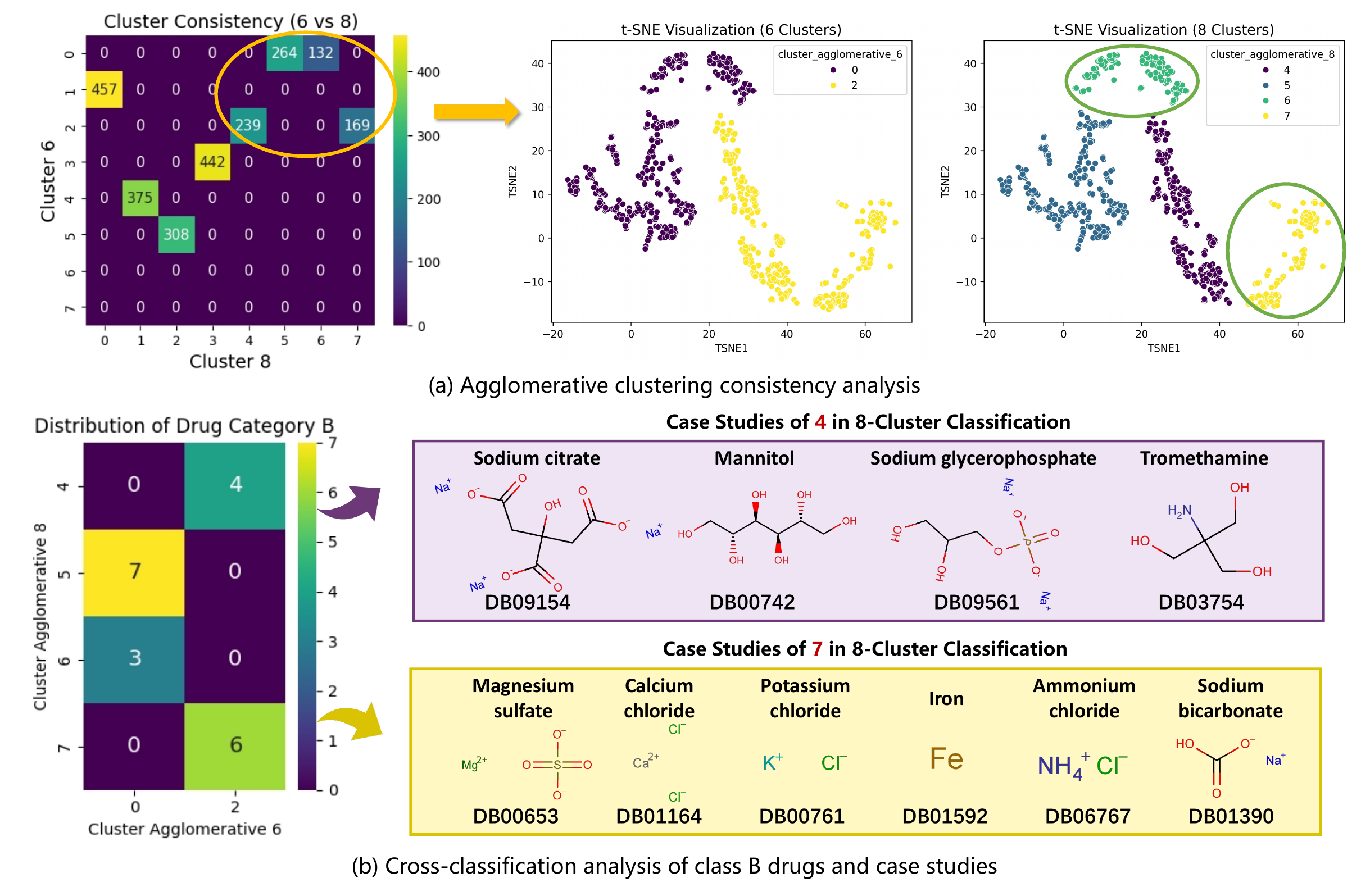}
\caption{\textbf{Cluster analysis and case studies.}
(a) Agglomerative clustering consistency analysis.
It compares the consistency of agglomerative clustering with 6 and 8 clusters, visualizing the clustering results of samples split into clusters 0 and 2 in the 6-cluster analysis to observe distribution differences.
(b) Cross-classification analysis of class B drugs and case studies.
The samples from clusters 0 and 2 in the 6-cluster analysis were selected, focusing on drugs from DrugBank's BLOOD AND BLOOD FORMING ORGANS category (B) to observe drug class distribution.
The ten molecules from cluster 2 in the 6-cluster analysis were chosen as case studies to analyze the basis and validity of the classification.}
\label{fig: case_study}
\end{figure*}

\subsection{Case studies}
To validate the pharmacological relevance of our clustering, we quantified the correspondence between t-SNE clustering labels and ATC Level-1 categories (e.g., `A' for Alimentary Tract and Metabolism, `V' for Various) using Trimmed Purity and KL Divergence metrics across various configurations (K-means, Birch, Agglomerative; 5--20 clusters).
As detailed in the Supplementary Information IV, the average Trimmed Purity across these configurations was approximately 0.90, indicating that 90\% of drugs were consistently assigned to clusters dominated by a single ATC category.
The average KL Divergence of 0.30 further confirms that ATC distributions within clusters are significantly more concentrated than the global distribution.
To analyze the practical significance of knowledge priors from a clustering perspective, we randomly selected the agglomerative clustering results with 6 and 8 clusters to observe distribution differences, as these two configurations exhibited significant variations.
As shown in Figure \ref{fig: case_study} (a), we first analyzed the clustering consistency and found that the main difference between the 6 and 8 clusters was the splitting of cluster 0 and cluster 1 in the 6-cluster result, which were divided into clusters 6 and 7 in the 8-cluster result.
We then visualized the split distributions.

The DrugBank database provides classifications for some drugs.
For example, sodium citrate has the ATC code B05CB02.
Here, we only consider the primary category indicated by the first letter of the ATC code, such as BLOOD AND BLOOD FORMING ORGANS (B).
Other categories include A - ALIMENTARY TRACT AND METABOLISM, G - GENITO URINARY SYSTEM AND SEX HORMONES, among others.

We further identified the primary categories of these drugs.
As shown in Figure \ref{fig: case_study} (b), we selected drugs with the primary category B.
We then chose ten molecules from cluster 2 in the 6-cluster result as case studies to analyze the reasonableness of the clustering method's classification.
The molecules in cluster 2 of the 6-cluster result have relatively simple structures, most of which include ions and exhibit electrolyte behavior, making this clustering reasonable.
The 8-cluster analysis split cluster 2 from the 6-cluster analysis into clusters 4 and 7.
The compounds classified as 7 are simple ionic compounds or elements, while those classified as 4 have more complex structures and functional groups.
This classification is also reasonable.
However, the 8-cluster experimental results were not entirely satisfactory, possibly due to the increased granularity leading to overfitting, especially on the `rare' data split.

\section{Discussion}
\label{sec:dis}

In our exploration, we have highlighted the adaptive prior knowledge framework for DDIE predictions.
Our methodology demonstrates the LLM's inherent capability to leverage drug knowledge.
This section aims to discuss the insights and limitations encountered throughout our study.
As shown in Figure \ref{fig: result}, incorporating prior drug knowledge significantly improves DDIE prediction, especially under limited data conditions.
However, as the amount of training data increases, the model tends to focus more on learning the intrinsic representations of drug molecules, gradually diminishing the impact of prior knowledge.

The ATC Codes \cite{nahler2009anatomical} (Anatomical Therapeutic Chemical Classification System) categorize drugs based on the organ or system they act upon and their therapeutic, pharmacological, and chemical properties.
While the DrugBank database provides classifications for many drugs, not all drug molecules have corresponding types.
We delve into the approach utilized for transforming drug features into drug types.
The drug features should be suitable for visualization and clustering.
We choose these features, as opposed to alternative representations, because of their appropriateness for quantitative analysis and compatibility with dimensionality reduction techniques like t-SNE.
While our study focuses on adaptive knowledge extraction and parameter optimization, the lack of systematic evaluation of alternative prompt designs is a limitation, as prompt engineering serves as an auxiliary tool.
This method is particularly favored for its effectiveness in preserving the local structure of the data while reducing dimensionality, making it easier to identify patterns and relationships inherent in the high-dimensional space.
We can explore diverse prompt structures to further enhance performance in future work.

There are multiple modalities of molecular representations, including one-dimensional data representations like fingerprints, SMILES, SELFIES, and descriptions.
Different knowledge combination strategies and model hyperparameters can lead to substantial performance variations.
The adaptive strategy selector based on RL can reduce the strategy space during the exploration process, thereby improving exploration efficiency.

Despite these advances, several practical limitations remain.
First, the benefits of prior knowledge tend to diminish as the volume of training data grows.
Second, the current RL-based selector, while efficient, may still converge to local optima.
Most importantly, when predicting DDIs involving novel drugs unseen during training, our framework currently requires re-clustering the entire drug corpus (including the new compounds) to assign the most effective input modality and knowledge-fusion strategy.
This process, though effective, introduces unnecessary computational overhead and hinders seamless deployment in real-world scenarios where new molecules are routinely encountered.

In future research, we will address the current limitations in several ways.
We will develop more sophisticated methods for integrating prior knowledge with molecular representations to sustain its contribution even as training data volume increases.
To enable efficient prediction on novel drugs, we will replace the offline re-clustering step with a lightweight end-to-end drug category classifier that directly predicts the optimal input modality and knowledge-fusion strategy for any unseen compound.
We will also enhance the adaptive strategy selector by incorporating advanced techniques, such as meta-learning and evolutionary algorithms, enabling it to escape local optima more effectively and discover globally superior configurations.
Additionally, we will explore the interpretability of our models, focusing on how prior knowledge and molecular representations influence model training, such as analyzing weight changes and attention distributions.
Furthermore, we plan to conduct a more systematic investigation of diverse prompt structures and their interactions with knowledge fusion to boost performance and generalization further.

\section{Conclusion}
\label{sec:con}
In this study, we propose a knowledge augmentation framework that infuses prior drug knowledge into LLMs, thereby enhancing the precision of DDIE predictions, particularly in few-shot scenarios.
This framework employs RL to adaptively refine knowledge extraction and synthesis, leveraging RL techniques to efficiently optimize the large strategy space and support the core contribution of knowledge integration.
This research introduces a novel paradigm for incorporating prior knowledge into LLMs, paving the way for more accurate DDIE predictions and other molecular science fields.

\vskip 6mm
\noindent







\vskip 2mm
\noindent
\textbf{Acknowledgment}
\vskip 2mm

\noindent
This work is supported by the Guangdong Science and Technology Programme (Grant No. 2024B0101010003).
This work is also supported by grants from the National Natural Science Foundation of China (62172456 and 62372484) and Pengcheng Cloud-Brain.

\vskip 2mm

\renewcommand\refname{\textbf{References}}

\bibliographystyle{IEEEtran}
\bibliography{references}

\begin{strip}
\end{strip}


\mbox{}
\clearpage
\clearpage

\end{document}